\documentclass[11pt]{article}

\usepackage[final]{acl}

\usepackage{times}
\usepackage{latexsym}

\usepackage[T1]{fontenc}

\usepackage[utf8]{inputenc}

\usepackage{microtype}

\usepackage{inconsolata}

\usepackage{graphicx}

\usepackage{soul}
\usepackage{url}
\usepackage{amssymb}
\usepackage{amsfonts}
\usepackage{xcolor}
\usepackage{booktabs}
\usepackage{amsmath}
\usepackage{amsthm}
\usepackage[switch]{lineno}
\usepackage{multirow}
\usepackage{float}
\usepackage{orcidlink}
\usepackage{adjustbox}
\usepackage[small,compact]{titlesec}
\usepackage{placeins} 
\usepackage{stfloats}
\newcommand{\squishlist}{
   \begin{list}{$\bullet$}
      { \setlength{\itemsep}{0pt} 
        \setlength{\parsep}{0pt} 
        \setlength{\topsep}{0pt} 
        \setlength{\partopsep}{0pt} 
        \setlength{\leftmargin}{1.5em} 
        \setlength{\labelwidth}{1em} 
        \setlength{\labelsep}{0.5em} } 
}

\newcommand{\FRorcid}{\orcidlink{0000-0002-1697-8586}}
\newcommand{\MEAorcid}{\orcidlink{0000-0002-4360-1362}}
\newcommand{\IMBorcid}{\orcidlink{0000-0002-7972-9085}}
\newcommand{\RPorcid}{\orcidlink{0000-0001-8059-5224}}
\newcommand{\MLorcid}{\orcidlink{0000-0001-5765-0416}}

\newcommand{\squishend}{\end{list}}
\title{The Alignment Bottleneck in Decomposition-Based Claim Verification}

\author{Mahmud Elahi Akhter\textsuperscript{1}\MEAorcid,
    Federico Ruggeri\textsuperscript{2}\FRorcid,
    Iman Munire Bilal\textsuperscript{3}\IMBorcid, \\
    \textbf{Rob Procter}\textsuperscript{3}\RPorcid,
    \textbf{Maria Liakata}\textsuperscript{1,4}\MLorcid \\
    \textsuperscript{1}Queen Mary University of London, UK,
    \textsuperscript{2}University of Bologna, Italy \\
    \textsuperscript{3}University of Warwick, UK,
    \textsuperscript{4}The Alan Turing Institute, UK \\
  \texttt{\{m.akhter, m.liakata\}@qmul.ac.uk}, \\
  \texttt{federico.ruggeri6@unibo.it},
  \texttt{\{iman.bilal, rob.procter\}@warwick.ac.uk} \\
  }

\begin{document}
\maketitle
\begin{abstract}
Structured claim decomposition is often proposed as a solution for verifying complex, multi-faceted claims, yet empirical results have been inconsistent. We argue that these inconsistencies stem from two overlooked bottlenecks: evidence alignment and sub-claim error profiles. To better understand these factors, we introduce a new dataset of real-world complex claims, featuring temporally bounded evidence and human-annotated sub-claim evidence spans. We evaluate decomposition under two evidence alignment setups: Sub-claim Aligned Evidence (SAE) and Repeated Claim-level Evidence (SRE). Our results reveal that decomposition brings significant performance improvement only when evidence is granular and strictly aligned. By contrast, standard setups that rely on repeated claim-level evidence (SRE) fail to improve and often degrade performance as shown across different datasets and domains (PHEMEPlus, MMM-Fact, COVID-Fact). Furthermore, we demonstrate that in the presence of noisy sub-claim labels, the nature of the error ends up determining downstream robustness. We find that conservative "abstention" significantly reduces error propagation compared to aggressive but incorrect predictions. These findings suggest that future claim decomposition frameworks must prioritize precise evidence synthesis and calibrate the label bias of sub-claim verification models.
\end{abstract}

\section{Introduction}

The task of \textit{claim verification} involves determining the \textit{veracity} for a given \textit{claim} based on associated \textit{evidence}. It is crucial for a number of different tasks, such as rumour or misinformation verification and fact-checking \citep{zubiaga-etal-2018-detection,glockner-etal-2022-missing,guo_michael_2022,pan-etal-2023-fact,Dmonte2024ClaimVI}. Claim verification is particularly challenging when dealing with complex claims which conflate multiple facts, requiring granular reasoning over evidence for accurate verification. To verify complex claims,  humans identify several sub-parts (sub-claims), gather multiple pieces of evidence and perform multi-step reasoning \citep{ijcai2021p619}. 

In such cases mapping the claim to a complete set of evidence is non-trivial; A common approach is to disentangle the sub-claims from the initial claim, (claim decomposition), which is then followed by sub-claim verification and 
rationale generation for each resulting sub-claim \citep{chen-etal-2022-generating,pan-etal-2023-fact}. Although laborious, this process offers more transparency in the decision-making process, especially in cases of insufficient or contradictory evidence relevant to different parts of a claim. However, this introduces an additional error propagation factor since errors in sub-claim decomposition or sub-claim labels can mislead claim-level inference.
\begin{figure*}[t]
\centering
\includegraphics[width=1.9\columnwidth]{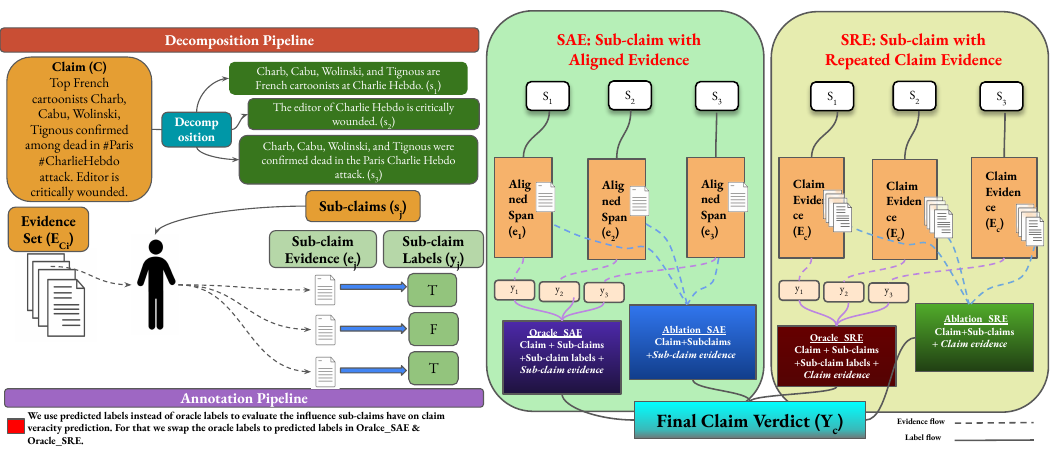}
\caption{\label{fig:pipelines} shows our annotation and claim verification pipeline and different setups for the study. Oracle\_(SAE/SRE) setups use gold sub-claim labels, ablation models do not use any sub-claim labels and noisy setup (not shown in figure) uses predicted sub-claim labels.}
\end{figure*}
Performing claim verification using Large Language Models (LLMs) \citep{wang-shu-2023-explainable,Quelle2023ThePA, Dmonte2024ClaimVI} has recently gained popularity. Apart from single LLMs, Agentic systems have also been applied to perform complex claim verification \citep{local_ma_2025,Liu2025BiDeVBD}. However, these still inherit the error propagation caveats due to decomposition and misalignment of granular evidence.  

Real world claim verification requires complex multi-hop reasoning steps. Reasoning with LLMs suffers from important shortcomings. For instance, ~\citet{dougrezlewis2025assessingreasoningcapabilitiesllms} show that LLMs struggle with abductive reasoning while LLM performance on claim verification is very much domain-dependent. Specifically, LLMs are better at handling simple claims from Wikipedia compared to real-world data which manifest  complex claims and hierarchical structure. This leads to a major bottleneck in claim verification and fact-checking since most datasets, where the state-of-the-art (SOTA) is trained and evaluated, are sourced from Wikipedia \citep{guo_michael_2022}.    

Studies that have evaluated the benefits of claim decomposition for fact-checking report conflicting results~\citep{tang-etal-2024-minicheck,wanner-etal-2024-closer,chen-etal-2024-complex}.
According to~\citet{tang-etal-2024-minicheck} claim decomposition is not necessary to improve LLM performance on fact checking. By contrast,~\citet{wanner-etal-2024-closer} show that the impact of decomposition is method-specific, especially when considering LLM-based decomposition. \citet{hu-etal-2024-decomposition-dilemma} carried out an in-depth analysis of claim decomposition, revealing several deciding factors, including how the decomposition process is defined and how evidence information is associated with each sub-claim. Thus, the benefits of claim decomposition for complex claims are currently poorly understood. We argue these conflicting findings largely stem from two factors: (i) how evidence is aligned and aggregated at the sub-claim level, and (ii) whether sub-claim veracity signals are reliable or noisy. Both of these can directly affect error propagation to claim-level inference. Based on this, we hypothesize that decomposition helps only when evidence is aligned at the sub-claim level, and sub-claim veracity labels are reliable; otherwise, decomposition can degrade performance in claim verification.

Here we investigate how sub-claim decomposition impacts complex claim verification in real-world data. Fig.~\ref{fig:pipelines} illustrates our approach. We assume that a complex claim \( C \) corresponding to an evidence set $E_C$ is decomposed into sub-claims \( \{ s_j \}_{j=1}^{m} \), each of which are paired with corresponding evidence sets \( \{ E_{j} \}_{j=1}^{m} \) and annotated with sub-claim veracity labels \( \{ y_j \}_{j=1}^{m} \). These structured triples \( (s_j, E_j, y_j) \) serve as inputs for sub-claim-level prediction and, when aggregated, form the basis for claim-level veracity inference under both oracle and noisy labels. 
Our approach connects human annotation, claim decomposition, and model evaluation into a unified framework for complex claim verification. We show that decomposition can improve performance under ideal labels (oracle), but can be neutral or harmful when labeling is noisy. Furthermore, to test the generalisability of the decomposition effect, we evaluate the same pipelines not only on temporally bounded rumour verification ( \citep{dougrez-lewis-etal-2022-phemeplus}), but also on two additional fact-checking datasets (MMM-Fact \citep{xu2025mmmfactmultimodalmultidomainfactchecking} and COVID-Fact\citep{saakyan-etal-2021-covid}), which differ with respect to domain and evidence characteristics.
We make the following contributions, 
\squishlist
\item We introduce a dataset of real-world claims decomposed into sub-claims accompanied by manually annotated sub-claim-level evidence and fine-grained veracity labels (\S\ref{dataset}).
    
\item  We formalize two complementary experimental settings (i) \textit{claim veracity} under oracle sub-claim labeling (upper bound) (\S\ref{sec:claim}) and under predicted sub-claim labeling (real-world scenario) (\S\ref{sec:noisy-sub-claims}),and (ii) \textit{sub-claim veracity} (\S\ref{sec:sub-claim}) as the upstream task; together these quantify how sub-claim labeling errors propagate to claim-level verification. 
    
\item We show decomposition improves performance under sub-claim aligned evidence and reliable sub-claim signals, but can also be neutral or harmful under noisy labeling and coarse evidence. Thus, we highlight that the benefits of decomposition are conditional, dependent on evidence quality and sub-claim error signals. 
\squishend
\section{Related work} \label{sec:related}
\noindent \textbf{Automatic Claim Decomposition.}
Several works have proposed decomposing fact-checking claims into atomic text segments. 
Such segments (sub-claims/atomic facts/atomic propositions)~\citep{dmonte-etal-2024-verification-llm-survey}, are typically extracted by prompting LLMs~\citep{zhang-gao-2023-towards}, via hybrid approaches relying on keywords~\citep{gong-etal-2024-key-claim-verification} or predefined templates~\citep{pan-etal-2023-fact,liu-etal-2024-teller}. Prompts are either textual propositions~\citep{kamoi-etal-2023-wice,tang-etal-2024-minicheck} or queries for information retrieval~\citep{chen-etal-2022-generating,chen-etal-2024-complex}. Apart from prompting methods, orchestrated multi-agent systems have also been used for decomposition-based claim verification \citep{zhao-etal-2024-pacar,local_ma_2025}.
~\citep{hu-etal-2024-decomposition-dilemma} provide valuable insights for evaluating the effects of claim decomposition. However, their work pertains to datasets containing automatically extracted sub-claim evidence and does not include real-world complex claims, unlike the task we address here.

\noindent \textbf{Claim Decomposition Datasets.}
Researchers have applied claim decomposition for several purposes, including fine-grained evidence retrieval~\citep{kamoi-etal-2023-wice,zhang-gao-2023-towards}, and factual verification of generated content to mitigate hallucinations~\citep{min-etal-2023-factscore,balepur-etal-2023-mastering-abcds-complex-questions,wang-etal-2024-factcheck,zhang-etal-2024-decomposition-benchmark, Gkoumas_Liakata_2024}. 
\begin{table}[t]
\small
\centering
\begin{tabular}{l|l|c|c|c}
\toprule
\textbf{Category} & \textbf{Split} & \multicolumn{1}{c}{\textbf{T\%}} & \multicolumn{1}{c}{\textbf{U\%}} & \multicolumn{1}{c}{\textbf{F\%}} \\
\toprule
\multirow{3}{*}{\textbf{Claim Veracity}} 
  & \textbf{Total} & 48.37 & 31.33 & 20.3 \\
  & \textbf{Train} & 47.98 & 31.78 & 20.25 \\
  & \textbf{Test}  & 50.00 & 29.49 & 20.51 \\
\midrule
\multirow{3}{*}{\textbf{Sub-claim Veracity}} 
  & \textbf{Total} & 57.66 & 34.05 & 8.3 \\
  & \textbf{Train} & 57.11 & 34.70 & 8.19 \\
  & \textbf{Test}  & 59.75 & 31.54 & 8.71 \\
\bottomrule
\end{tabular}
\caption{Combined distribution of claim veracity and sub-claim labels across the full dataset, training, and test splits, shown as percentage.}
\label{tab:no_fold}
\end{table}

To this effect, several fact-checking datasets containing sub-claims have been proposed. Notable examples are BingCheck, for evaluating the actuality of LLM-generated texts in response to fact-checking~\citep{li-etal-2024-self}, LLM-AggreFact~\citep{tang-etal-2024-minicheck} containing a suite of fact-checking corpora, ClaimDecomp centered on extracting explicit and implicit sub-claims~\citep{chen-etal-2022-generating}, WiCE for fine-grained evidence retrieval on Wikipedia~\citep{kamoi-etal-2023-wice}, and AVERITEC for real-world claim verification~\citep{schlichtkrull-etal-2023-averitec}.

ClaimDecomp~\citep{chen-etal-2022-generating} and WiCE~\citep{kamoi-etal-2023-wice} are the closest to our setting. In ClaimDecomp, sub-claims are presented as questions to help evaluate the veracity of complex claims extracted from Politifact.
Both the claim and the evidence-based justification, written by journalists, are used to reverse-engineer the verification process, thus introducing both literal (produced from the claim) and implied sub-claims (produced from evidence). 
Here we adhere to a more realistic setting of decomposing claims with no prior knowledge of existing evidence. In WiCE, claim decomposition is carried out via LLMs, and each sub-claim is associated with evidence manually annotated from Wikipedia articles.
Similarly, our dataset provides manually annotated evidence texts for sub-claims but unlike WiCE, we focus on real-world claim verification, consisting of social media content, where rumours often emerge and spread~\citep{zubiaga-etal-2018-detection}.

\section{Dataset}
\label{dataset}
\subsection{Task Overview}
Our task involves assessing the veracity of complex claims through structured decomposition into sub-claims, each supported by relevant evidence. 
We first decompose claims into distinct sub-claims, then independently verify each sub-claim using associated evidence. 
We then analyze how this pipeline behaves under different evidence association strategies (claim-level vs sub-claim-aligned evidence) and different supervision regimes (oracle vs predicted sub-claim labels), to quantify error propagation to claim-level inference.

\subsection{Dataset creation}
\label{sec:method}
We filter complex claims within the PHEME dataset \citep{zubiaga-etal-2016-pheme} by removing claims containing fewer than three verbs and two sentences. This heuristic is a proxy method for multi-fact composition introduced by \citet{chen-etal-2022-generating}. PHEME contains a collection of claims posted on Twitter as time-sensitive breaking news events were unfolding within a specific time-frame. It contains 1972 rumourous claims related to five events and each of the rumours are annotated with a veracity label (True, False, Unverified). To decompose complex claims into sub-claims, we adopted a fixed decomposition procedure to isolate the effects of evidence association and label noise on verification performance, rather than optimizing the decomposition model itself. We used an LLM-based schema introduced by FactScore \citep{min-etal-2023-factscore} with gpt-3.5-turbo-instruct to break down text into short statements covering only one piece of information.  This is followed by a manual evaluation process using the claim detection guidelines in \citet{sundriyal-etal-2022-empowering} to ensure that the resulting sub-claims contain check-worthy information and are comprehensive when considered without the original claim. In total, we extract 423 complex claims from PHEME, covering five distinct events. We then leveraged PHEMEplus \citep{dougrez-lewis-etal-2022-phemeplus} annotations to map these claims to relevant evidence from news articles. PHEMEplus contains evidence-providing articles for the claims within PHEME in a time constrained manner (within 2 days of each event). We obtained 399 claims after accounting for evidence quality and dataset artifacts in PHEMEplus. We filtered out image-only evidence, hyperlinks and embedded ads. We do not consider claims with associated images in the tweet since decomposing such claims result in non-sense sub-claims (URL links or pictures). 
\paragraph{Domain generalisation.} We additionally use MMM-Fact and COVID-Fact to test generalisability of our findings beyond event-centric rumour verification. MMM-Fact is a large-scale multi-domain fact-checking benchmark with multi-source evidence. It has a full size of 125,449 samples with evidence grouped into retrieval-difficulty levels (Basic/Intermediate/Advanced). We use intermediate level which had 6-10 evidence sources per claim (21,873). COVID-Fact is a domain-specific (health/COVID-19) real-world fact-checking dataset with 4,086 claims paired with single source evidence. Their goal was to test multi-document synthesis (MMM-Fact), to better assess evidence alignment and to see the decomposition and alignment effect in a narrower-domain with less structural complexity (COVID-Fact).
\subsection{Annotation details}
Given a claim, associated sub-claims and evidence providing articles for the original claim, the task was to label the veracity of the sub-claims and highlight relevant spans of evidence supporting the assigned veracity label. The highlighted evidence was later extracted as sub-claim level evidence. The annotator guidelines can be found in Appendix~\ref{app:guide}. 
In particular, each sub-claim was labeled as true (T), false (F), and Unverified (U). Starting with 1169 sub-claims, these were evenly distributed to three annotators, Phd students in Computer Science, fluent in English, so that 300 samples were annotated by all. The latter were used to calculate the Inter Annotator Agreement (IAA) score. In short, we double-annotated 300 sub-claims for IAA; the remaining sub-claims were single-annotated and distributed evenly, resulting in $\approx$290 unique items per annotator. Annotators labeled a sub-claim as (T) if directly supported by the evidence, (F) if contradicted, and (U) if the evidence was insufficient or ambiguous. Evidence spans were selected as the minimal set of sentences justifying the label. 
We computed IAA as Bennett’s S score to account for label imbalance, resulting in a score of 0.81. We also analyzed sentence-level overlap among annotators using BLEU and BERTScore. The Average BLEU\citep{papineni-etal-2002-bleu} ranged from $\approx$ 0.40–0.48 across pairs, and average symmetric BERTScore \citep{zhang2020bertscoreevaluatingtextgeneration} F1 was around 0.49–0.56, indicating moderate to strong overlap in selected evidence spans. 

Unlike Averitec, \citep{Schlichtkrull2023AVeriTeCAD}, our verification process was not constructed as a QA process but rather the annotators had to verify sub-claims based on the provided evidence. Contrary to Factcheck-Bench's \citep{wang-etal-2024-factcheck} automatic evidence retrieval, we manually selected evidence spans to isolate the verification problem from retrieval errors\citep{Ge2025ResolvingCE}.

Our dataset differs from previous work in that we provide: 
(i) human-annotated evidence texts per extracted sub-claim rather than evidence automatically extracted by LLMs and (ii) evidence documents per claim with a closed retrieval setting.
In particular, we consider only evidence documents that were restricted to articles published within a fixed temporal window around the event \citep{dougrez-lewis-etal-2022-phemeplus}, preventing leakage from post-hoc reporting and ensuring a realistic real-time verification setting \citep{zubiaga-etal-2018-detection}. Furthermore, this controlled evidence setting allows us to isolate how sub-claim evidence alignment and sub-claim label noise affect decomposition-based verification, without conflating effects from open-domain retrieval.

\subsection{Dataset details}
\label{sec:details}
Dataset statistics are in Table~\ref{tab:no_fold}, including distribution of veracity labels for both complex claims and sub-claims, with 399 complex claims 
corresponding to 1169 sub-claims.  We provide an example of claim verification through sub-claims in Appendix~\ref{app:exp}. We used the full dataset for oracle and ablation setups in Tables~\ref{tab:f1_balacc_comparison_all_datasets_paired} and ~\ref{tab:ablation_and_noisy_claim_level} with 929 samples for training the baseline and GNN models in \S\ref{sec:models} and 240 as testing. The results in Tables ~\ref{tab:ablation_and_noisy_claim_level}, ~\ref{tab:sub-claim_f1_summary}, and  ~\ref{tab:subclaim_profile}, were calculated using this test set. 
On MMM-Fact we extracted 1181 (5.40\% of intermediate subset) complex claims and 163 complex claims in COVID-Fact (3.99\% full dataset). These resulted in 2326 and 447 sub-claims respectively for each dataset. Note that these datasets do not contain manually annotated sub-claim labels or corresponding sub-claim aligned evidence.

\section{Experiments} 
\label{sec:exp}
While claim verification has increasingly been addressed alongside explanation generation \citep{atanasova-etal-2020-generating-fact,info13100500,bilal2024generatingzeroshotabstractiveexplanations}, here we focus on assessing the merits of claim decomposition given (i) claim-level verification  and (ii) sub-claim verification labels, while considering two evidence alignment settings \textbf{(SRE Eq.~\eqref{eq:sre} vs SAE Eq.~\eqref{eq:sae})} (\emph{(\S\ref{sec:sub-claim})}) and two levels of labeling quality for sub-claims (oracle vs predicted/noisy). Evidence is temporally bounded and pre-linked (no open-domain retrieval) to isolate decomposition effects from open retrieval errors.  

We denote as \( C \) a complex claim and as \(E_C=\{E_{Ci}\}_{i=1}^{n}\) the complete evidence set corresponding to the original claim \( C \). \(C\) is decomposed into a set of \( m \) sub-claims \( \{ s_j \}_{j=1}^{m} \), each associated with an evidence set \( \{ E_j \}_{j=1}^{m} \) and a veracity label \( y_j \in \{T, F, U\} \). Here, \( y_j\) denotes either oracle sub-claim labels (upper bound), predicted labels (realistic), or is omitted in label-free ablation variants (Eq.~\eqref{eq:abl_sre} \& Eq.~\eqref{eq:abl_sae}).  
The overall claim-level veracity \( Y_C \) will depend on the veracity and evidence of its sub-claims as follows:
\begin{equation}
    Y_C = f(C,\{(s_j, E_j, y_j)\}_{j=1}^{m}) 
\end{equation}
where each sub-claim evidence \(E_j\) derives from some claim evidence \(E_{Ci}\).
To ensure temporal and contextual fidelity, evidence sets are restricted to information available before the claim timestamp:
\begin{equation}
    E_{Ci} \subseteq \{ e \in \mathcal{D} : t(e) \leq t(C) \}
\end{equation}
where \( \mathcal{D} \) is the document collection, \( t(e) \) is the publication time of evidence \( e \), and \( t(C) \) is the time when the claim \( C \) emerged.  
This constraint prevents open-domain retrieval to avoid compromising the time-sensitive nature of the dataset as it would otherwise retrieve evidence not available at the time of unfolding events; events happening in real time usually generate unverified claims that are later resolved within a few hours to a few days. As our goal was not to propose a new retrieval method, we fix evidence to measure the causal effect of decomposition choices instead of compounding it with retrieval errors. 
\subsection{Claim Veracity}
\label{sec:claim_veracity}
\paragraph{Experiment details.}
Here the task is to verify a claim (label as T/F) given a set of evidence texts, known to be relevant to the claim where evidence documents are pre-linked to each claim via PHEMEPLUS within a temporally bounded window.
We do not consider unverified (U) labels and drop them as these were not typically supported by corresponding evidence within the PHEMEPlus subset.  
All experiments at claim-level are conducted using the Qwen3-14B \citep{yang2025qwen3technicalreport} reasoning model.

We first perform claim-level veracity prediction using \textbf{oracle sub-claim labels} \citep{huang2024largelanguagemodelsselfcorrect,yang-etal-2024-improving} to establish an upper bound on performance. Here, $y_j$ the gold sub-claim is provided as part of the structured input to test the best-case scenario for decomposition; these are later replaced with predicted labels Eq.~\eqref{eq:sub_pred} or omitted in ablations (Eq.~\eqref{eq:abl_sre} \& Eq.~\eqref{eq:abl_sae}) to reflect realistic usage.  
The model predicts claim veracity as:
\begin{equation} \label{eq:1}
\hat{Y}_C = \mathcal{M}(C, \mathcal{E})    
\end{equation}
where \( \mathcal{E} \) represents the evidence configuration used.  
We consider the following evidence alignment configurations per setting: Claim-level evidence ($E_{Ci}$), for the Vanilla and \textbf{Sub-claims with Repeated Claim-level Evidence (SRE)} experimental settings, and Sub-claim level evidence ($E_j$) for the  \textbf{Sub-claims with Aligned Evidence (SAE)} setting: 

\begin{align}
    \textbf{(i) Vanilla:} &\quad \mathcal{E} = E_{C}=\{E_{Ci}\}_{i=1}^{n} \label{eq:vanilla} \\
    \textbf{(ii) SRE:} &\quad \mathcal{E} = (s_j,\{E_{Ci}\}_{i=1}^{n},y_j)_{j=1}^{m} \label{eq:sre} \\
    \textbf{(iii) SAE:} &\quad \mathcal{E} = \{(s_j, E_j, y_j)\}_{j=1}^{m} \label{eq:sae}
\end{align}

In practice, SRE repeats the same claim-level evidence block after each sub-claim in the prompt, whereas SAE inserts the sub-claim-specific evidence block $E_j$ next to its sub-claim $s_j$.
\paragraph{Ablation.}
We conduct a set of targeted ablation studies, with the aim of isolating the impact of the oracle sub-claim veracity labels on downstream model performance and quantifying the influence of evidence alignment. Here, we use the same model setting as Equation \ref{eq:1}, but in the ablations the evidence for SRE and SAE isn't paired with any sub-claim veracity labels.
\begin{align}
    \textbf{(i) Abl. SRE:} &\quad \mathcal{E} = (s_j,\{E_{Ci}\}_{i=1}^{n})_{j=1}^{m} \label{eq:abl_sre} \\
    \textbf{(ii) Abl. SAE:} &\quad \mathcal{E} = \{(s_j, E_j)\}_{j=1}^{m} \label{eq:abl_sae}
\end{align}
Ablation SRE also mirrors real-world use cases where we may have sub-claims available, with the parent claim and corresponding evidence but no sub-claim labels or sub-claim level evidence. Indeed, Ablation SRE is the setup used in the experiments for the MMM-Fact and COVID-Fact datasets. The SAE setting was not possible for these as contrary to the dataset introduced in this paper, they lack human-annotated sub-claim evidence spans, a resource-intensive annotation process.
\paragraph{Metrics and statistical significance.} Claim-level verification performance (See Table~\ref{tab:f1_balacc_comparison_all_datasets_paired}) is gauged using two primary \emph{\textbf{metrics:}} macro-averaged F1-score and balanced accuracy \citep{5597285}. We use both due to the class imbalance between claim veracity labels. 
Models were run over three independent random seeds (six for ablations).

\noindent \emph{\textbf{Statistical Significance:}} We applied paired statistical testing on claims across model configurations to evaluate the statistical significance of our findings. We performed a paired bootstrap analysis (Primary Performance $\Delta$) to assess the differences in F1 and Accuracy metrics. Claims were selected with replacement ($N=1000$ resamples) to compute the performance difference ($\Delta = \text{Metric}_{\text{SAE/SRE}} - \text{Metric}_{\text{baseline}}$) for each resample. We report the two-sided bootstrap $p$-value ($p_{\text{boot}}$) (how often $\Delta$s crosses zero). To evaluate the significance of changes in correct/incorrect predictions, we utilized McNemar’s test (Categorical Agreement) on the $2 \times 2$ table of paired correctness on the same claims. This focuses specifically on instances where (SAE/SRE)) and baseline disagree. We report both the McNemar $p$-value and the Odds Ratio (OR) of the disagreeing pairs ($b_{01}/b_{10}$) as a measure of effect size, indicating how many times more likely the model is to correct a baseline error than to introduce a new one.

\subsection{Sub-claim Veracity}
\noindent \textbf{Experiment details:} Sub-claim veracity is formulated as a classification task \citep{guo_michael_2022} where individual sub-claims are classified with the claim-level evidence (claim level evidence is repeated per sub-claim). Each sub-claim is considered to be a discrete instance. 
At the sub-claim level, each sub-claim \( s_j \) is independently classified as:
\begin{equation}
\hat{y}_j = \mathcal{M}(s_j,\{E_{Ci}\}_{i=1}^{n})
\end{equation}
We predict $\hat{y}_j \in {T,F,U}$, since decomposed sub-claims can remain unresolved even when the overall claim is verifiable. We evaluate using the macro-averaged F1-score. We do not consider sub-claim veracity prediction given sub-claim evidence as this deviates from real-world scenarios where only claim-level evidence is available e.g. to journalists.

Predicted sub-claim veracity labels are then aggregated to infer claim-level veracity under noisy sub-claim labels (setting SRE\_Noisy in Table~\ref{tab:ablation_and_noisy_claim_level}):
\begin{equation}
\hat{Y}_C^{(\text{noisy})} = f(C,\{(s_j, \mathcal{E}, \hat{y}_j)\}_{j=1}^{m}) \label{eq:sub_pred}
\end{equation}
Here, \(f\) is the classifier that aggregates the claim and sub-claim information. This setup allows us to assess how noisy sub-claim predictions \(\hat{y}_j)\) influence overall claim verification. We use an LLM-based aggregator, and leave a systematic comparison against deterministic or learned aggregation rules (e.g., conjunctive/disjunctive logic over $\hat{y}_j$, or meta-classifiers) as future work.
\subsection{Models for sub-claim veracity prediction}
\label{sec:models}
\noindent \textbf{Graph Neural Network Baseline.}
We chose GNNs as a strong structured baseline. Our Graph Neural Network (GNN) is a variant of \citep{bilal2024generatingzeroshotabstractiveexplanations}, able to leverage evidence from both news articles and Twitter threads, tested on PHEMEplus. It is structured as a bipartite graph to perform sub-claim veracity prediction 
with two distinct node sets: tweet nodes (preserving thread structure) ($\mathbf{x}_t \in \mathbb{R}^{n \times 512}$) and evidence nodes ($\mathbf{x}_e \in \mathbb{R}^{m \times 128}$), interconnected via dedicated edges. Tweet and evidence texts are encoded into initial node embeddings using pretrained encoders. The training was done with leave-one-event-out \citep{kochkina-liakata-2020-estimating} (learning rate=$5e^{-5}$, weight decay= $1e^{-4}$, batch size=$20$). More details provided in Appendix~\ref{app:gnn}.\\
\noindent \textbf{Encoder Models.}
We use CHEF \citep{hu-etal-2022-chef} a BERT-based sub-claim verifier that performs evidence selection before veracity classification.
\begin{table*}[t]
\centering
\small
\setlength{\tabcolsep}{1.5pt}
\renewcommand{\arraystretch}{1.15}
\begin{adjustbox}{width=\textwidth}
\begin{tabular}{ll rrrr cccc}
\toprule
\textbf{Setup} & \textbf{Dataset} & \textbf{F1 $\pm$ std} & \textbf{$\Delta$} & \textbf{$p_{\text{boot}}$} & \textbf{OR / McNemar($p$)} & \textbf{Balanced Acc. $\pm$ std} & \textbf{$\Delta$} & \textbf{$p_{\text{boot}}$} & \textbf{OR / McNemar($p$)} \\
\midrule

Vanilla &
\multirow{3}{*}{PHEMEplus} &
$0.5643 \pm 0.0091$ & -- & -- & -- &
$0.6072 \pm 0.0074$ & -- & -- & -- \\
Oracle\_SRE & &
$0.5872 \pm 0.0127$ & +0.0229 & 0.124 & 1.52 / 0.0227 &
$0.6117 \pm 0.0136$ & +0.0045 & 0.774 & 1.52 / 0.0227 \\
Oracle\_SAE & &
\textbf{0.6268 $\pm$ 0.0098} & \textbf{+0.0625} & \textbf{0.0000} & \textbf{2.13 / 0.0000} &
\textbf{0.6558 $\pm$ 0.0135} & \textbf{+0.0486} & \textbf{0.0000} & \textbf{2.13 / 0.0000} \\
\midrule

\textbf{Vanilla} &
\multirow{2}{*}{CovidFact} &
\textbf{0.7365 $\pm$ 0.0248} & -- & -- & -- &
\textbf{0.7559 $\pm$ 0.0240} & -- & -- & -- \\
Ablation\_SRE & &
$0.7252 \pm 0.0161$ & -0.0113 & 0.549 & 0.96 / 0.5599 &
$0.7533 \pm 0.0148$ & -0.0326 & 0.115 & 0.86 / 0.1153 \\
\midrule

\textbf{Vanilla} &
\multirow{2}{*}{MMM-Fact} &
\textbf{0.7550 $\pm$ 0.0069} & -- & -- & -- &
\textbf{0.7451 $\pm$ 0.0065} & -- & -- & -- \\
Ablation\_SRE & &
$0.6878 \pm 0.0064$ & -0.0672 & 0.0000 & 0.05 / 0.0000 &
$0.7120 \pm 0.0127$ & -0.0377 & 0.0000 & 0.05 / 0.0000 \\
\bottomrule
\end{tabular}
\end{adjustbox}
\caption{Claim-level veracity given oracle sub-claim labels (PHEMEPlus) and no sub-claim labels (CovidFact, MMM-Fact)\S(\ref{sec:details}), to assess the effect of evidence alignment. Comparison of Macro F1 and Balanced Accuracy with \textbf{paired} statistical analysis relative to the base model. $\Delta$ shows absolute performance change vs.\ Vanilla. Paired bootstrap $p$-values reports the two-sided probability of $\Delta$ crossing zero. `OR / McNemar($p$)' report the paired odds ratio (OR) and McNemar($p$) for paired correctness differences. Best results per dataset/metric in bold.}
\label{tab:f1_balacc_comparison_all_datasets_paired}
\end{table*}
We report the two strongest CHEF variants: \textit{CHEF$_{latent}$}
(hard selection via Hard Kumaraswamy~\citep{bastings-etal-2019-interpretable}) and \textit{CHEF$_{rl}$}(Bernoulli selection optimized with RL~\citep{lei-etal-2016-rationalizing}).
We additionally use BERT\citep{devlin-etal-2019-bert} as a strong encoder baseline: larger encoders tended to overfit (low-resource setting) in our preliminary trials, so we use BERT as a competitive and standard control. We chunk long evidence to fit the model context (learning rate=$5e^{-5}$, weight decay=$1e^{-4}$, batch size=$8$).
\\\noindent \textbf{Large Language Model.}
For both claim and sub-claim veracity prediction our LLM of choice is Qwen3-14B reasoning model. We evaluate Qwen3-14B in zero-shot mode with a fixed prompt (temperature=$0.3$). The prompts are detailed in Appendix~\ref{app:prompt}. We initially used Llama 3.1 8B-instruct and Phi4 Reasoning-14B  but found both failed at structured input based prediction. We discuss this more in Appendix~\ref{app:llm_models}. 
\section{Results}
\label{sec:results}
\subsection{Claim Veracity Results}
\label{sec:claim}
As Table~\ref{tab:f1_balacc_comparison_all_datasets_paired} shows, the benefits of decomposition depend heavily on how the evidence is aligned. On PHEMEPlus, decomposition offers a clear advantage when sub-claim evidence is available and aligned with sub-claims (SAE).
 We see statistically significant gains 
over the Vanilla baseline (+ 0.0625 F1 and +0.0486 balanced accuracy). However, simply repeating claim-level evidence (SRE) on the same dataset, without associated sub-claim evidence to match the sub-claims and their labels, produces negligible changes, suggesting that usefulness of decomposition is conditional upon evidence alignment.
The limitations of SRE become even clearer on MMM-Fact and COVID-Fact where it fails to improve verification on either dataset. Vanilla slightly edges out on COVID-Fact, while on MMM-Fact, the drop from providing noisy sub-claims (without labels or evidence) is substantial (macro-F1 $\Delta:-0.0672$, balanced accuracy $\Delta:-0.0521$). These findings validate our core hypothesis that decomposition works when evidence is tightly coupled with valid sub-claim signals. Without such alignment, the process tends to introduce noise that harms the final inference, a phenomenon we analyze further in (\S\ref{sec:noisy-sub-claims}). These findings highlight the need for proper evidence synthesis. 
\subsection{Ablation results}
\label{sec:ablation}
Table~\ref{tab:ablation_and_noisy_claim_level} highlights a trade-off between evidence granularity and the quality of sub-claim labels. When using redundant claim-level evidence but no sub-claim veracity labels (Ablation SRE), the model relies little on sub-claim veracity. However, in the aligned setting where sub-claim evidence is provided (Ablation SAE), the absence of sub-claim veracity labels is detrimental. The model performs significantly worse than standard SAE and even falls below the SRE baseline. This indicates that granular, sub-claim-level, evidence is insufficient on its own. Without the signal from correct sub-claim veracity labels, the model struggles to synthesize the fine-grained information, leading to inference errors that coarser evidence setups seem to avoid.
\begin{table*}[t]
\small
\centering
\begin{tabular}{lcccc}
\toprule
\textbf{Setup} & \textbf{F1 Macro $\pm$ std} & $\Delta$ \textbf{F1} & \textbf{Balanced Accuracy $\pm$ std} & $\Delta$ \textbf{Acc} \\
\midrule
Oracle\_SRE           & 0.5872 $\pm$ 0.0127          & --        & 0.6117 $\pm$ 0.0136                           & --        \\   
Ablation\_SRE         & 0.5808 $\pm$ 0.0064          & $-0.0064$ & 0.6259 $\pm$ 0.0052          & $+0.0142$  \\
\midrule
Oracle\_SAE           & 0.6268 $\pm$ 0.0098          & --        & 0.6558 $\pm$ 0.0132                          & --        \\
Ablation\_SAE         & 0.5485 $\pm$ 0.0052          & $-0.0783$ & 0.6220 $\pm$ 0.0065          & $-0.0338$  \\
\midrule
\multicolumn{5}{l}{\textbf{Effect of Noisy sub-claim Predictions on Claim Verification (all $\Delta$ vs. SAE\_oracle)}} \\
\midrule
\textbf{Oracle\_SAE} & \textbf{0.6268 $\pm$ 0.0098} & --        & \textbf{0.6558 $\pm$ 0.0132} & --         \\
Qwen Noisy\_SAE       & 0.5964 $\pm$ 0.0360          & $-0.0304$ & 0.6425 $\pm$ 0.0595          & $-0.0133$  \\
Qwen Noisy\_SRE       & 0.4335 $\pm$ 0.0439          & $-0.1933$ & 0.4411 $\pm$ 0.0512          & $-0.2147$  \\
GNN Noisy\_SAE        & 0.5839 $\pm$ 0.1202          & $-0.0429$ & 0.5963 $\pm$ 0.1361          & $-0.0595$  \\
GNN Noisy\_SRE        & 0.4416 $\pm$ 0.0489          & $-0.1852$ & 0.4399 $\pm$ 0.0609          & $-0.2159$  \\
\bottomrule
\end{tabular}
\caption{Claim-level veracity performance comparison under ablations and noisy sub-claim prediction inputs for PhemePLUS (\S\ref{sec:details}). Ablation $\Delta$s are computed relative to the corresponding full model with sub-claim labels. That is, Ablation\_SRE (no sub-claim labels, claim-level evidence) vs. Oracle\_SRE (oracle sub-claim labels, claim-level evidence) and  Ablation\_SAE (no sub-claim  labels, sub-claim level evidence) vs. Oracle\_SAE (oracle sub-claim labels, sub-claim-level evidence). In the noisy setting, $\Delta$s are computed relative to \textbf{Oracle\_SAE}, where noisy variants use predicted sub-claim labels and oracle uses the gold sub-claim labels. Noisy labels were obtained either via GNN or Qwen3-14B.}
\label{tab:ablation_and_noisy_claim_level}
\end{table*}

\begin{table}[t]
\centering
\small
\begin{tabular}{lcc}
\toprule
\textbf{Model} & \textbf{F1 Macro ($\%$)} \\
\midrule
CHEF$_{latent}$ & $27.30 \pm 7.11$ \\
CHEF$_{rl}$     & $28.92 \pm 6.56$ \\
Bert-Chunk    & $41.90 \pm 0.57$ \\
GNN    & $45.79 \pm 1.25$ \\
Qwen3-14B & \textbf{$56.94 \pm 0.92$} \\
\bottomrule
\end{tabular}
\caption{Macro F1 scores ($\%$) with standard deviation across 3 seeds for sub-claim-level veracity classification on PhemePLUS (\S\ref{sec:details}).}
\label{tab:sub-claim_f1_summary}
\end{table}
\subsection{Sub-claim Veracity Results}
\label{sec:sub-claim}
Table~\ref{tab:sub-claim_f1_summary} presents the sub-claim veracity results. The zero-shot Qwen3-14B sets the state of the art with $56.94\%$ macro-F1. The performance gap between the GNN ($45.79\%$) and the encoder-based CHEF variants ($\approx 28\%$) underscores the necessity of structured modeling for this task. While encoders struggle with overfitting and evidence integration, the GNN successfully captures evidence interactions, though it still falls short of the LLM's reasoning capabilities. However, the low-resource nature of the dataset makes it much harder for trained encoders to perform well compared to much larger LLMs.
\subsection{Effect of Noisy sub-claim Predictions on Claim Verification}
\label{sec:noisy-sub-claims}
Table~\ref{tab:ablation_and_noisy_claim_level} shows that replacing oracle sub-claim labels with predicted ones reduces performance. However, the severity of this drop depends on the evidence configuration. Under the aligned SAE setting, the degradation is moderate (Qwen SAE\textsubscript{Noisy}: $\Delta-0.0304$ F1). However, under SRE, the models collapse (dropping to $\approx 0.44$ F1), confirming that redundant claim-level evidence exacerbates error propagation from noisy sub-claim labels. This error propagation pattern aligns with findings from \citet{lyu2025mitigating} in multi-hop verification, confirming that intermediate step errors compound through the pipeline. 
A deeper look at the per-model error analysis reveals distinct model biases. From Table~\ref{tab:subclaim_profile} we can see that, during sub-claim prediction, when GNN commits (non-\textit{U}), its accuracy on T/F (Acc$_{v}$) is comparable to Qwen (0.625 vs 0.646), but it abstains far more (35\% vs 15.8\%) and almost never detects refutations ($R_F$ 0.039 vs.\ 0.314). This gap is driven primarily by coverage on verifiable sub-claims (T/F). Qwen issues a polarity label for 85.7\% of verified cases (Cov$_{ver}$=0.857), while GNN commits on only 66.7\% (0.667). Moreover, Qwen’s increased refutation sensitivity comes with modest precision ($P_F$ 0.276 vs.\ 0.154), indicating more frequent false positives for F alongside higher $R_F$. Thus, decomposition success depends not only on correctness when committing, but on the model’s commit/abstain and refute/affirm bias. This distinction explains the downstream effects where in the presence of fine-grained evidence (SAE), the GNN's tendency to predict \textit{Unverified} acts as a safety mechanism, but also misses crucial refutations. Ultimately, this analysis shows that sub-claim macro-F1 is a poor predictor of success and that error analysis, especially the balance between conservative abstention and aggressive detection determines whether decomposition can improve verification. 
\begin{table}[t]
\centering
\small
\setlength{\tabcolsep}{2pt}
\begin{tabular}{lccccccc}
\toprule
\textbf{Model} & \%U & \%F & $R_F$ & $P_F$ & Cov$_{ver}$ & Acc$_{{v|!u}}$ & Acc$_{v}$\\
\midrule
Qwen & 15.8 & 24.2 & 0.314 & 0.276 & 0.857 & 0.554 & 0.646\\
GNN  & 35.0 & 5.4  & 0.039 & 0.154 & 0.667 & 0.417 & 0.625\\
\bottomrule
\end{tabular}
\caption{Sub-claim-labeling profile (F/U) (\S\ref{sec:details}). T prediction very similar, at 60 and 59.6\% respectively for Qwen and GNN. $R_F,P_F$ are refutation recall/precision.
Cov$_{ver}$ is non-abstain rate on verified sub-claims (GT$\in\{T,F\}$). Acc$_{v|!u}$ counts abstentions as errors.}
\label{tab:subclaim_profile}
\end{table}
\section{Discussion and Future work}
\label{sec:disc}
\noindent \textbf{The Alignment Bottleneck.} Our experiments clarify a boundary condition for decomposition-based verification: claim decomposition is not a universal performance booster, but rather depends on evidence granularity. Decomposition improves performance only when sub-claims are paired with sub-claim-aligned evidence (SAE setting). When evidence is repeated at the claim level (SRE), decomposition adds structure without resolving ambiguity and can degrade performance as seen on MMM-Fact and COVID-Fact, where the SRE setting failed to improve over Vanilla. This suggests that the primary real-world bottleneck is often \textit{evidence synthesis}. Without reliable sub-claim–evidence alignment, decomposition offers limited value and may rather amplify noise.

\noindent \textbf{Sensitivity to Error Profiles.} Moving from oracle to predicted sub-claim labels shows upstream noise affects decomposition, but the \textit{type} of labeling error matters more than overall accuracy. Under noise, SRE collapses, while SAE can remain comparatively stable depending on the predictor’s label bias. We observe that the GNN behaves conservatively, frequently predicting \textit{Unverified}; in a granular-evidence setting this can be beneficial, by avoiding incorrect polar signals. By contrast, Qwen is biased towards refutations and risks more false positives. Thus,stabilizing decomposition pipelines may require optimizing the sub-claim predictor’s \emph{label bias} rather than maximizing macro-F1 alone.
\noindent \textbf{Future Directions: Human-Centric and Real-Time Verification.} Beyond algorithmic accuracy, the ultimate utility of decomposition lies in its interpretability. A key avenue for future work is evaluating whether sub-claim-level rationales genuinely assist human fact-checkers (e.g., journalists) in identifying missing evidence or isolating disputed claims, rather than simply increasing cognitive load. Finally, our current setup uses temporally bounded but static evidence snapshots. In breaking news scenarios, such as those on social media, information evolves rapidly. Future systems must move towards \textit{just-in-time} verification, utilizing timeline summarization to capture evolving narratives and explicitly representing the status of sub-claims as "unresolved" until sufficient evidence becomes available.

\section{Conclusion}
\label{sec:conclusion}
We investigate the role of structured sub-claim decomposition in complex claim verification. By leveraging a real-world dataset with human-annotated sub-claim evidence spans, we show that decomposition is \emph{not} a universal remedy: it requires granular, sub-claim-aligned evidence and reliable veracity labels to be effective. However, \textbf{our results reveal that when these ideal conditions are met, sub-claim decomposition leads to significant performance gains over vanilla methods that treat claims monolithically.}

By contrast, when evidence is limited to the claim level or when sub-claim labels are noisy, the benefits of decomposition diminish or even become detrimental to performance, as shown with our findings on MMM-Fact and Covid-Fact. Our ablation experiments further confirm a critical interaction between evidence granularity and label reliability: providing granular evidence without reliable labels can confuse the verification model. Furthermore, downstream robustness depends heavily on the \emph{label bias} of the verification model (e.g., the safety of abstaining via an \textit{Unverified} label versus the risk of committing to incorrect polarity). Ultimately, we conclude that the primary bottleneck for deployment in real-world settings is not decomposition itself, but the capability for precise evidence synthesis and alignment.

Future research should prioritize developing robust sub-claim predictors that optimize for label bias, specifically favoring calibrated abstention over incorrect polarity. Additionally, critical advancements are needed in evidence synthesis to produce sub-claim-aligned spans, particularly under temporal constraints. Ultimately, the goal is to design verifiers that can leverage structured decomposition while remaining resilient to noisy labeling inherent in real-world settings.

\section*{Limitations}

\paragraph{Closed evidence setting and temporal window assumptions.}
To isolate the causal effects of decomposition choices, we fix evidence to a temporally bounded, pre-linked snapshot and avoid open-domain retrieval; this removes retrieval noise but also limits realism for end-to-end systems where evidence must be retrieved and updated over time. Additionally, our evidence is temporally constrained around unfolding events, which may not match settings with longer verification horizons. 

\paragraph{Label-space mismatch between claim- and sub-claim-level tasks.}
At claim level we restrict prediction to \{T,F\} (dropping \textit{Unverified}), while sub-claim verification explicitly models \{T,F,U\}; this mismatch can obscure how abstention should propagate to the final claim verdict and may understate the value of calibrated uncertainty in real deployments.

\paragraph{Model and prompt sensitivity; limited end-to-end validation.}
Our claim-level results are based on a specific long-context LLM setup (Qwen3-14B) and fixed prompting templates; other models may differ substantially in structured-input adherence and bias profiles, affecting downstream robustness under noise. We also use a fixed decomposition procedure to avoid confounding factors, so we do not evaluate learned decomposition or retrieval components end-to-end. 

\section*{Ethics Statement}
Both PHEME and PHEMEPlus are pre-existing datasets of rumours, for the development of which ethical approval was obtained by the original research team. Our annotations add sub-claims and associated evidence without the addition of any ethical issues.

\section*{Acknowledgments}
This work was supported by a UKRI/EPSRC Turing AI Fellowship to Maria Liakata (grant ref EP/V030302/1) and the Alan Turing Institute (grant ref EP/N510129/1). This work was also supported by the Engineering and Physical Sciences Research Council [grant number EP/Y009800/1], through funding from Responsible Ai UK (KP0016) as a Keystone project lead by Maria Liakata. F. Ruggeri is partially supported by the project European Commission's NextGeneration EU programme, PNRR -- M4C2 -- Investimento 1.3, Partenariato Esteso, PE00000013 - ``FAIR - Future Artificial Intelligence Research'' -- Spoke 8 ``Pervasive AI’’ and by the European Union’s Justice Programme under Grant Agreement No. 101087342 for the project “Principles Of Law In National and European VAT”. We also thank Queen Mary University of London Comp-Research Team for their infrastructure support.

\bibliography{custom}

@article{lyu2025mitigating,
  title={Mitigating Error Propagation in Multi-hop Fact Verification with Logic Reasoning},
  author={Lyu, Xiucheng and Cao, Chengyu and Sun, Mingwei and Xu, Ruifeng},
  journal={International Journal of Machine Learning and Cybernetics},
  year={2025}
}

@misc{zhang2020bertscoreevaluatingtextgeneration,
      title={BERTScore: Evaluating Text Generation with BERT}, 
      author={Tianyi Zhang and Varsha Kishore and Felix Wu and Kilian Q. Weinberger and Yoav Artzi},
      year={2020},
      eprint={1904.09675},
      archivePrefix={arXiv},
      primaryClass={cs.CL},
      url={https://arxiv.org/abs/1904.09675}, 
}

@inproceedings{papineni-etal-2002-bleu,
    title = "{B}leu: a Method for Automatic Evaluation of Machine Translation",
    author = "Papineni, Kishore  and
      Roukos, Salim  and
      Ward, Todd  and
      Zhu, Wei-Jing",
    editor = "Isabelle, Pierre  and
      Charniak, Eugene  and
      Lin, Dekang",
    booktitle = "Proceedings of the 40th Annual Meeting of the Association for Computational Linguistics",
    month = jul,
    year = "2002",
    address = "Philadelphia, Pennsylvania, USA",
    publisher = "Association for Computational Linguistics",
    url = "https://aclanthology.org/P02-1040/",
    doi = "10.3115/1073083.1073135",
    pages = "311--318"
}

@inproceedings{saakyan-etal-2021-covid,
    title = "{COVID}-Fact: Fact Extraction and Verification of Real-World Claims on {COVID}-19 Pandemic",
    author = "Saakyan, Arkadiy  and
      Chakrabarty, Tuhin  and
      Muresan, Smaranda",
    editor = "Zong, Chengqing  and
      Xia, Fei  and
      Li, Wenjie  and
      Navigli, Roberto",
    booktitle = "Proceedings of the 59th Annual Meeting of the Association for Computational Linguistics and the 11th International Joint Conference on Natural Language Processing (Volume 1: Long Papers)",
    month = aug,
    year = "2021",
    address = "Online",
    publisher = "Association for Computational Linguistics",
    url = "https://aclanthology.org/2021.acl-long.165/",
    doi = "10.18653/v1/2021.acl-long.165",
    pages = "2116--2129"
}

@misc{xu2025mmmfactmultimodalmultidomainfactchecking,
      title={MMM-Fact: A Multimodal, Multi-Domain Fact-Checking Dataset with Multi-Level Retrieval Difficulty}, 
      author={Wenyan Xu and Dawei Xiang and Tianqi Ding and Weihai Lu},
      year={2025},
      eprint={2510.25120},
      archivePrefix={arXiv},
      primaryClass={cs.SI},
      url={https://arxiv.org/abs/2510.25120}, 
}

@inproceedings{devlin-etal-2019-bert,
    title = "{BERT}: Pre-training of Deep Bidirectional Transformers for Language Understanding",
    author = "Devlin, Jacob  and
      Chang, Ming-Wei  and
      Lee, Kenton  and
      Toutanova, Kristina",
    editor = "Burstein, Jill  and
      Doran, Christy  and
      Solorio, Thamar",
    booktitle = "Proceedings of the 2019 Conference of the North {A}merican Chapter of the Association for Computational Linguistics: Human Language Technologies, Volume 1 (Long and Short Papers)",
    month = jun,
    year = "2019",
    address = "Minneapolis, Minnesota",
    publisher = "Association for Computational Linguistics",
    url = "https://aclanthology.org/N19-1423/",
    doi = "10.18653/v1/N19-1423",
    pages = "4171--4186"
}

@misc{huang2024largelanguagemodelsselfcorrect,
      title={Large Language Models Cannot Self-Correct Reasoning Yet}, 
      author={Jie Huang and Xinyun Chen and Swaroop Mishra and Huaixiu Steven Zheng and Adams Wei Yu and Xinying Song and Denny Zhou},
      year={2024},
      eprint={2310.01798},
      archivePrefix={arXiv},
      primaryClass={cs.CL},
      url={https://arxiv.org/abs/2310.01798}, 
}

@article{yang-etal-2024-improving,
    title = "Improving Probability-based Prompt Selection Through Unified Evaluation and Analysis",
    author = "Yang, Sohee  and
      Kim, Jonghyeon  and
      Jang, Joel  and
      Ye, Seonghyeon  and
      Lee, Hyunji  and
      Seo, Minjoon",
    journal = "Transactions of the Association for Computational Linguistics",
    volume = "12",
    year = "2024",
    address = "Cambridge, MA",
    publisher = "MIT Press",
    url = "https://aclanthology.org/2024.tacl-1.37/",
    doi = "10.1162/tacl_a_00666",
    pages = "664--680"
}

@article{Ge2025ResolvingCE,
  title={Resolving Conflicting Evidence in Automated Fact-Checking: A Study on Retrieval-Augmented LLMs},
  author={Ziyu Ge and Yuhao Wu and Daniel Wai Kit Chin and Roy Ka-Wei Lee and Rui Cao},
  journal={ArXiv},
  year={2025},
  volume={abs/2505.17762},
  url={https://api.semanticscholar.org/CorpusID:278886039}
}

@article{Schlichtkrull2023AVeriTeCAD,
  title={AVeriTeC: A Dataset for Real-world Claim Verification with Evidence from the Web},
  author={M. Schlichtkrull and Zhijiang Guo and Andreas Vlachos},
  journal={ArXiv},
  year={2023},
  volume={abs/2305.13117},
  url={https://api.semanticscholar.org/CorpusID:258832336}
}

@inproceedings{lei-etal-2016-rationalizing,
    title = "Rationalizing Neural Predictions",
    author = "Lei, Tao  and
      Barzilay, Regina  and
      Jaakkola, Tommi",
    editor = "Su, Jian  and
      Duh, Kevin  and
      Carreras, Xavier",
    booktitle = "Proceedings of the 2016 Conference on Empirical Methods in Natural Language Processing",
    month = nov,
    year = "2016",
    address = "Austin, Texas",
    publisher = "Association for Computational Linguistics",
    url = "https://aclanthology.org/D16-1011",
    doi = "10.18653/v1/D16-1011",
    pages = "107--117",
}

@inproceedings{bastings-etal-2019-interpretable,
    title = "Interpretable Neural Predictions with Differentiable Binary Variables",
    author = "Bastings, Jasmijn  and
      Aziz, Wilker  and
      Titov, Ivan",
    editor = "Korhonen, Anna  and
      Traum, David  and
      M{\`a}rquez, Llu{\'\i}s",
    booktitle = "Proceedings of the 57th Annual Meeting of the Association for Computational Linguistics",
    month = jul,
    year = "2019",
    address = "Florence, Italy",
    publisher = "Association for Computational Linguistics",
    url = "https://aclanthology.org/P19-1284",
    doi = "10.18653/v1/P19-1284",
    pages = "2963--2977",
}

@inproceedings{hu-etal-2022-chef,
    title = "{CHEF}: A Pilot {C}hinese Dataset for Evidence-Based Fact-Checking",
    author = "Hu, Xuming  and
      Guo, Zhijiang  and
      Wu, GuanYu  and
      Liu, Aiwei  and
      Wen, Lijie  and
      Yu, Philip",
    editor = "Carpuat, Marine  and
      de Marneffe, Marie-Catherine  and
      Meza Ruiz, Ivan Vladimir",
    booktitle = "Proceedings of the 2022 Conference of the North American Chapter of the Association for Computational Linguistics: Human Language Technologies",
    month = jul,
    year = "2022",
    address = "Seattle, United States",
    publisher = "Association for Computational Linguistics",
    url = "https://aclanthology.org/2022.naacl-main.246",
    doi = "10.18653/v1/2022.naacl-main.246",
    pages = "3362--3376",
}

@inproceedings{chen-etal-2024-complex,
    title = "Complex Claim Verification with Evidence Retrieved in the Wild",
    author = "Chen, Jifan  and
      Kim, Grace  and
      Sriram, Aniruddh  and
      Durrett, Greg  and
      Choi, Eunsol",
    editor = "Duh, Kevin  and
      Gomez, Helena  and
      Bethard, Steven",
    booktitle = "Proceedings of the 2024 Conference of the North American Chapter of the Association for Computational Linguistics: Human Language Technologies (Volume 1: Long Papers)",
    month = jun,
    year = "2024",
    address = "Mexico City, Mexico",
    publisher = "Association for Computational Linguistics",
    url = "https://aclanthology.org/2024.naacl-long.196",
    doi = "10.18653/v1/2024.naacl-long.196",
    pages = "3569--3587",
}

@inproceedings{tang-etal-2024-minicheck,
    title = "{M}ini{C}heck: Efficient Fact-Checking of {LLM}s on Grounding Documents",
    author = "Tang, Liyan  and
      Laban, Philippe  and
      Durrett, Greg",
    editor = "Al-Onaizan, Yaser  and
      Bansal, Mohit  and
      Chen, Yun-Nung",
    booktitle = "Proceedings of the 2024 Conference on Empirical Methods in Natural Language Processing",
    month = nov,
    year = "2024",
    address = "Miami, Florida, USA",
    publisher = "Association for Computational Linguistics",
    url = "https://aclanthology.org/2024.emnlp-main.499",
    pages = "8818--8847",
}

@inproceedings{kamoi-etal-2023-wice,
    title = "{W}i{CE}: Real-World Entailment for Claims in {W}ikipedia",
    author = "Kamoi, Ryo  and
      Goyal, Tanya  and
      Diego Rodriguez, Juan  and
      Durrett, Greg",
    editor = "Bouamor, Houda  and
      Pino, Juan  and
      Bali, Kalika",
    booktitle = "Proceedings of the 2023 Conference on Empirical Methods in Natural Language Processing",
    month = dec,
    year = "2023",
    address = "Singapore",
    publisher = "Association for Computational Linguistics",
    url = "https://aclanthology.org/2023.emnlp-main.470",
    doi = "10.18653/v1/2023.emnlp-main.470",
    pages = "7561--7583",
}

@inproceedings{wang-etal-2024-factcheck,
    title = "Factcheck-Bench: Fine-Grained Evaluation Benchmark for Automatic Fact-checkers",
    author = "Wang, Yuxia  and
      Gangi Reddy, Revanth  and
      Mujahid, Zain Muhammad  and
      Arora, Arnav  and
      Rubashevskii, Aleksandr  and
      Geng, Jiahui  and
      Mohammed Afzal, Osama  and
      Pan, Liangming  and
      Borenstein, Nadav  and
      Pillai, Aditya  and
      Augenstein, Isabelle  and
      Gurevych, Iryna  and
      Nakov, Preslav",
    editor = "Al-Onaizan, Yaser  and
      Bansal, Mohit  and
      Chen, Yun-Nung",
    booktitle = "Findings of the Association for Computational Linguistics: EMNLP 2024",
    month = nov,
    year = "2024",
    address = "Miami, Florida, USA",
    publisher = "Association for Computational Linguistics",
    url = "https://aclanthology.org/2024.findings-emnlp.830",
    pages = "14199--14230",
}

@misc{dmonte-etal-2024-verification-llm-survey,
      title={Claim Verification in the Age of Large Language Models: A Survey}, 
      author={Alphaeus Dmonte and Roland Oruche and Marcos Zampieri and Prasad Calyam and Isabelle Augenstein},
      year={2024},
      eprint={2408.14317},
      archivePrefix={arXiv},
      primaryClass={cs.CL},
      url={https://arxiv.org/abs/2408.14317}, 
}

@inproceedings{li-etal-2024-self,
    title = "Self-Checker: Plug-and-Play Modules for Fact-Checking with Large Language Models",
    author = "Li, Miaoran  and
      Peng, Baolin  and
      Galley, Michel  and
      Gao, Jianfeng  and
      Zhang, Zhu",
    editor = "Duh, Kevin  and
      Gomez, Helena  and
      Bethard, Steven",
    booktitle = "Findings of the Association for Computational Linguistics: NAACL 2024",
    month = jun,
    year = "2024",
    address = "Mexico City, Mexico",
    publisher = "Association for Computational Linguistics",
    url = "https://aclanthology.org/2024.findings-naacl.12",
    doi = "10.18653/v1/2024.findings-naacl.12",
    pages = "163--181",
}

@inproceedings{schlichtkrull-etal-2023-averitec,
  author       = {Michael Schlichtkrull and
                  Zhijiang Guo and
                  Andreas Vlachos},
  editor       = {Alice Oh and
                  Tristan Naumann and
                  Amir Globerson and
                  Kate Saenko and
                  Moritz Hardt and
                  Sergey Levine},
  title        = {AVeriTeC: {A} Dataset for Real-world Claim Verification with Evidence
                  from the Web},
  booktitle    = {Advances in Neural Information Processing Systems 36: Annual Conference
                  on Neural Information Processing Systems 2023, NeurIPS 2023, New Orleans,
                  LA, USA, December 10 - 16, 2023},
  year         = {2023},
  url          = {http://papers.nips.cc/paper\_files/paper/2023/hash/cd86a30526cd1aff61d6f89f107634e4-Abstract-Datasets\_and\_Benchmarks.html},
  bibsource    = {dblp computer science bibliography, https://dblp.org}
}

@inproceedings{zhao-etal-2024-pacar,
    title = "{PACAR}: Automated Fact-Checking with Planning and Customized Action Reasoning Using Large Language Models",
    author = "Zhao, Xiaoyan  and
      Wang, Lingzhi  and
      Wang, Zhanghao  and
      Cheng, Hong  and
      Zhang, Rui  and
      Wong, Kam-Fai",
    editor = "Calzolari, Nicoletta  and
      Kan, Min-Yen  and
      Hoste, Veronique  and
      Lenci, Alessandro  and
      Sakti, Sakriani  and
      Xue, Nianwen",
    booktitle = "Proceedings of the 2024 Joint International Conference on Computational Linguistics, Language Resources and Evaluation (LREC-COLING 2024)",
    month = may,
    year = "2024",
    address = "Torino, Italia",
    publisher = "ELRA and ICCL",
    url = "https://aclanthology.org/2024.lrec-main.1099",
    pages = "12564--12573",
}

@article{gong-etal-2024-key-claim-verification,
  author       = {Haisong Gong and
                  Huanhuan Ma and
                  Qiang Liu and
                  Shu Wu and
                  Liang Wang},
  title        = {Navigating the Noisy Crowd: Finding Key Information for Claim Verification},
  journal      = {CoRR},
  volume       = {abs/2407.12425},
  year         = {2024},
  url          = {https://doi.org/10.48550/arXiv.2407.12425},
  doi          = {10.48550/ARXIV.2407.12425},
  eprinttype    = {arXiv},
  eprint       = {2407.12425},
  bibsource    = {dblp computer science bibliography, https://dblp.org}
}

@inproceedings{pan-etal-2023-fact,
    title = "Fact-Checking Complex Claims with Program-Guided Reasoning",
    author = "Pan, Liangming  and
      Wu, Xiaobao  and
      Lu, Xinyuan  and
      Luu, Anh Tuan  and
      Wang, William Yang  and
      Kan, Min-Yen  and
      Nakov, Preslav",
    editor = "Rogers, Anna  and
      Boyd-Graber, Jordan  and
      Okazaki, Naoaki",
    booktitle = "Proceedings of the 61st Annual Meeting of the Association for Computational Linguistics (Volume 1: Long Papers)",
    month = jul,
    year = "2023",
    address = "Toronto, Canada",
    publisher = "Association for Computational Linguistics",
    url = "https://aclanthology.org/2023.acl-long.386",
    doi = "10.18653/v1/2023.acl-long.386",
    pages = "6981--7004",
}

@inproceedings{liu-etal-2024-teller,
    title = "{TELLER}: A Trustworthy Framework for Explainable, Generalizable and Controllable Fake News Detection",
    author = "Liu, Hui  and
      Wang, Wenya  and
      Li, Haoru  and
      Li, Haoliang",
    editor = "Ku, Lun-Wei  and
      Martins, Andre  and
      Srikumar, Vivek",
    booktitle = "Findings of the Association for Computational Linguistics: ACL 2024",
    month = aug,
    year = "2024",
    address = "Bangkok, Thailand",
    publisher = "Association for Computational Linguistics",
    url = "https://aclanthology.org/2024.findings-acl.919",
    doi = "10.18653/v1/2024.findings-acl.919",
    pages = "15556--15583",
}

@misc{hu-etal-2024-decomposition-dilemma,
      title={Decomposition Dilemmas: Does Claim Decomposition Boost or Burden Fact-Checking Performance?}, 
      author={Qisheng Hu and Quanyu Long and Wenya Wang},
      year={2024},
      eprint={2411.02400},
      archivePrefix={arXiv},
      primaryClass={cs.IR},
      url={https://arxiv.org/abs/2411.02400}, 
}

@inproceedings{wanner-etal-2024-closer,
    title = "A Closer Look at Claim Decomposition",
    author = "Wanner, Miriam  and
      Ebner, Seth  and
      Jiang, Zhengping  and
      Dredze, Mark  and
      Van Durme, Benjamin",
    editor = "Bollegala, Danushka  and
      Shwartz, Vered",
    booktitle = "Proceedings of the 13th Joint Conference on Lexical and Computational Semantics (*SEM 2024)",
    month = jun,
    year = "2024",
    address = "Mexico City, Mexico",
    publisher = "Association for Computational Linguistics",
    url = "https://aclanthology.org/2024.starsem-1.13",
    doi = "10.18653/v1/2024.starsem-1.13",
    pages = "153--175",
}

@misc{zhang-etal-2024-decomposition-benchmark,
      title={A Claim Decomposition Benchmark for Long-form Answer Verification}, 
      author={Zhihao Zhang and Yixing Fan and Ruqing Zhang and Jiafeng Guo},
      year={2024},
      eprint={2410.12558},
      archivePrefix={arXiv},
      primaryClass={cs.CL},
      url={https://arxiv.org/abs/2410.12558}, 
}

@misc{balepur-etal-2023-mastering-abcds-complex-questions,
      title={Mastering the ABCDs of Complex Questions: Answer-Based Claim Decomposition for Fine-grained Self-Evaluation}, 
      author={Nishant Balepur and Jie Huang and Samraj Moorjani and Hari Sundaram and Kevin Chen-Chuan Chang},
      year={2023},
      eprint={2305.14750},
      archivePrefix={arXiv},
      primaryClass={cs.CL},
      url={https://arxiv.org/abs/2305.14750}, 
}

@inproceedings{zhang-gao-2023-towards,
    title = "Towards {LLM}-based Fact Verification on News Claims with a Hierarchical Step-by-Step Prompting Method",
    author = "Zhang, Xuan  and
      Gao, Wei",
    editor = "Park, Jong C.  and
      Arase, Yuki  and
      Hu, Baotian  and
      Lu, Wei  and
      Wijaya, Derry  and
      Purwarianti, Ayu  and
      Krisnadhi, Adila Alfa",
    booktitle = "Proceedings of the 13th International Joint Conference on Natural Language Processing and the 3rd Conference of the Asia-Pacific Chapter of the Association for Computational Linguistics (Volume 1: Long Papers)",
    month = nov,
    year = "2023",
    address = "Nusa Dua, Bali",
    publisher = "Association for Computational Linguistics",
    url = "https://aclanthology.org/2023.ijcnlp-main.64",
    doi = "10.18653/v1/2023.ijcnlp-main.64",
    pages = "996--1011",
}

@article{zubiaga-etal-2018-detection,
author = {Zubiaga, Arkaitz and Aker, Ahmet and Bontcheva, Kalina and Liakata, Maria and Procter, Rob},
title = {Detection and Resolution of Rumours in Social Media: A Survey},
year = {2018},
issue_date = {March 2019},
publisher = {Association for Computing Machinery},
address = {New York, NY, USA},
volume = {51},
number = {2},
issn = {0360-0300},
url = {https://doi.org/10.1145/3161603},
doi = {10.1145/3161603},
journal = {ACM Comput. Surv.},
month = feb,
articleno = {32},
numpages = {36}
}

@inproceedings{sundriyal-etal-2022-empowering,
    title = "Empowering the Fact-checkers! Automatic Identification of Claim Spans on {T}witter",
    author = "Sundriyal, Megha  and
      Kulkarni, Atharva  and
      Pulastya, Vaibhav  and
      Akhtar, Md. Shad  and
      Chakraborty, Tanmoy",
    editor = "Goldberg, Yoav  and
      Kozareva, Zornitsa  and
      Zhang, Yue",
    booktitle = "Proceedings of the 2022 Conference on Empirical Methods in Natural Language Processing",
    month = dec,
    year = "2022",
    address = "Abu Dhabi, United Arab Emirates",
    publisher = "Association for Computational Linguistics",
    url = "https://aclanthology.org/2022.emnlp-main.525",
    doi = "10.18653/v1/2022.emnlp-main.525",
    pages = "7701--7715"
}

@inproceedings{chen-etal-2022-generating,
    title = "Generating Literal and Implied Subquestions to Fact-check Complex Claims",
    author = "Chen, Jifan  and
      Sriram, Aniruddh  and
      Choi, Eunsol  and
      Durrett, Greg",
    editor = "Goldberg, Yoav  and
      Kozareva, Zornitsa  and
      Zhang, Yue",
    booktitle = "Proceedings of the 2022 Conference on Empirical Methods in Natural Language Processing",
    month = dec,
    year = "2022",
    address = "Abu Dhabi, United Arab Emirates",
    publisher = "Association for Computational Linguistics",
    url = "https://aclanthology.org/2022.emnlp-main.229",
    doi = "10.18653/v1/2022.emnlp-main.229",
    pages = "3495--3516",
}

@inproceedings{min-etal-2023-factscore,
    title = "{FA}ct{S}core: Fine-grained Atomic Evaluation of Factual Precision in Long Form Text Generation",
    author = "Min, Sewon  and
      Krishna, Kalpesh  and
      Lyu, Xinxi  and
      Lewis, Mike  and
      Yih, Wen-tau  and
      Koh, Pang  and
      Iyyer, Mohit  and
      Zettlemoyer, Luke  and
      Hajishirzi, Hannaneh",
    editor = "Bouamor, Houda  and
      Pino, Juan  and
      Bali, Kalika",
    booktitle = "Proceedings of the 2023 Conference on Empirical Methods in Natural Language Processing",
    month = dec,
    year = "2023",
    address = "Singapore",
    publisher = "Association for Computational Linguistics",
    url = "https://aclanthology.org/2023.emnlp-main.741",
    doi = "10.18653/v1/2023.emnlp-main.741",
    pages = "12076--12100",
    
}

@inproceedings{dougrez-lewis-etal-2022-phemeplus,
    title = "{PHEMEP}lus: Enriching Social Media Rumour Verification with External Evidence",
    author = "Dougrez-Lewis, John  and
      Kochkina, Elena  and
      Arana-Catania, Miguel  and
      Liakata, Maria  and
      He, Yulan",
    editor = "Aly, Rami  and
      Christodoulopoulos, Christos  and
      Cocarascu, Oana  and
      Guo, Zhijiang  and
      Mittal, Arpit  and
      Schlichtkrull, Michael  and
      Thorne, James  and
      Vlachos, Andreas",
    booktitle = "Proceedings of the Fifth Fact Extraction and VERification Workshop (FEVER)",
    month = may,
    year = "2022",
    address = "Dublin, Ireland",
    publisher = "Association for Computational Linguistics",
    url = "https://aclanthology.org/2022.fever-1.6",
    doi = "10.18653/v1/2022.fever-1.6",
    pages = "49--58",
}

@inproceedings{glockner-etal-2022-missing,
    title = "Missing Counter-Evidence Renders {NLP} Fact-Checking Unrealistic for Misinformation",
    author = "Glockner, Max  and
      Hou, Yufang  and
      Gurevych, Iryna",
    editor = "Goldberg, Yoav  and
      Kozareva, Zornitsa  and
      Zhang, Yue",
    booktitle = "Proceedings of the 2022 Conference on Empirical Methods in Natural Language Processing",
    month = dec,
    year = "2022",
    address = "Abu Dhabi, United Arab Emirates",
    publisher = "Association for Computational Linguistics",
    url = "https://aclanthology.org/2022.emnlp-main.397/",
    doi = "10.18653/v1/2022.emnlp-main.397",
    pages = "5916--5936"
}

@article{guo_michael_2022,
    author = {Guo, Zhijiang and Schlichtkrull, Michael and Vlachos, Andreas},
    title = {A Survey on Automated Fact-Checking},
    journal = {Transactions of the Association for Computational Linguistics},
    volume = {10},
    pages = {178-206},
    year = {2022},
    month = {02},
    issn = {2307-387X},
    doi = {10.1162/tacl_a_00454},
    url = {https://doi.org/10.1162/tacl\_a\_00454},
    eprint = {https://direct.mit.edu/tacl/article-pdf/doi/10.1162/tacl\_a\_00454/1987018/tacl\_a\_00454.pdf},
}

@inproceedings{local_ma_2025,
author = {Ma, Jiatong and Hu, Linmei and Li, Rang and Fu, Wenbo},
title = {LoCal: Logical and Causal Fact-Checking with LLM-Based Multi-Agents},
year = {2025},
isbn = {9798400712746},
publisher = {Association for Computing Machinery},
address = {New York, NY, USA},
url = {https://doi.org/10.1145/3696410.3714748},
doi = {10.1145/3696410.3714748},
booktitle = {Proceedings of the ACM on Web Conference 2025},
pages = {1614–1625},
numpages = {12},
location = {Sydney NSW, Australia},
series = {WWW '25}
}

@article{Dmonte2024ClaimVI,
  title={Claim Verification in the Age of Large Language Models: A Survey},
  author={Alphaeus Eric Dmonte and Roland Oruche and Marcos Zampieri and Prasad Calyam and Isabelle Augenstein},
  journal={ArXiv},
  year={2024},
  volume={abs/2408.14317},
  url={https://api.semanticscholar.org/CorpusID:271957454}
}

@article{Quelle2023ThePA,
  title={The perils and promises of fact-checking with large language models},
  author={Dorian Quelle and Alexandre Bovet},
  journal={Frontiers in Artificial Intelligence},
  year={2023},
  volume={7},
  url={https://api.semanticscholar.org/CorpusID:264406187}
}

@inproceedings{wang-shu-2023-explainable,
    title = "Explainable Claim Verification via Knowledge-Grounded Reasoning with Large Language Models",
    author = "Wang, Haoran  and
      Shu, Kai",
    editor = "Bouamor, Houda  and
      Pino, Juan  and
      Bali, Kalika",
    booktitle = "Findings of the Association for Computational Linguistics: EMNLP 2023",
    month = dec,
    year = "2023",
    address = "Singapore",
    publisher = "Association for Computational Linguistics",
    url = "https://aclanthology.org/2023.findings-emnlp.416/",
    doi = "10.18653/v1/2023.findings-emnlp.416",
    pages = "6288--6304"
}

@article{Liu2025BiDeVBD,
  title={BiDeV: Bilateral Defusing Verification for Complex Claim Fact-Checking},
  author={Yuxuan Liu and Hongda Sun and Wenya Guo and Xinyan Xiao and Cunli Mao and Zhengtao Yu and Rui Yan},
  journal={ArXiv},
  year={2025},
  volume={abs/2502.16181},
  url={https://api.semanticscholar.org/CorpusID:276575112}
}

@article{Gkoumas_Liakata_2024, title={Less for More: Enhanced Feedback-aligned Mixed LLMs for Molecule Caption Generation and Fine-Grained NLI  Evaluation}, DOI={10.26434/chemrxiv-2024-p39s0-v2}, journal={ChemRxiv}, author={Gkoumas, Dimitris and Liakata, Maria}, year={2024}}

@inproceedings{ijcai2021p619,
  title     = {Automated Fact-Checking for Assisting Human Fact-Checkers},
  author    = {Nakov, Preslav and Corney, David and Hasanain, Maram and Alam, Firoj and Elsayed, Tamer and Barrón-Cedeño, Alberto and Papotti, Paolo and Shaar, Shaden and Da San Martino, Giovanni},
  booktitle = {Proceedings of the Thirtieth International Joint Conference on
               Artificial Intelligence, {IJCAI-21}},
  publisher = {International Joint Conferences on Artificial Intelligence Organization},
  editor    = {Zhi-Hua Zhou},
  pages     = {4551--4558},
  year      = {2021},
  month     = {8},
  note      = {Survey Track},
  doi       = {10.24963/ijcai.2021/619},
  url       = {https://doi.org/10.24963/ijcai.2021/619},
}

@misc{dougrezlewis2025assessingreasoningcapabilitiesllms,
      title={Assessing the Reasoning Capabilities of LLMs in the context of Evidence-based Claim Verification}, 
      author={John Dougrez-Lewis and Mahmud Elahi Akhter and Federico Ruggeri and Sebastian Löbbers and Yulan He and Maria Liakata},
      year={2025},
      eprint={2402.10735},
      archivePrefix={arXiv},
      primaryClass={cs.CL},
      url={https://arxiv.org/abs/2402.10735}, 
}

@inproceedings{kochkina-liakata-2020-estimating,
    title = "Estimating predictive uncertainty for rumour verification models",
    author = "Kochkina, Elena  and
      Liakata, Maria",
    booktitle = "Proceedings of the 58th Annual Meeting of the Association for Computational Linguistics",
    month = jul,
    year = "2020",
    address = "Online",
    publisher = "Association for Computational Linguistics",
    url = "https://aclanthology.org/2020.acl-main.623",
    doi = "10.18653/v1/2020.acl-main.623",
    pages = "6964--6981",
}

@misc{yang2025qwen3technicalreport,
      title={Qwen3 Technical Report}, 
      author={An Yang and Anfeng Li and Baosong Yang and Beichen Zhang and Binyuan Hui and Bo Zheng and Bowen Yu and Chang Gao and Chengen Huang and Chenxu Lv and Chujie Zheng and Dayiheng Liu and Fan Zhou and Fei Huang and Feng Hu and Hao Ge and Haoran Wei and Huan Lin and Jialong Tang and Jian Yang and Jianhong Tu and Jianwei Zhang and Jianxin Yang and Jiaxi Yang and Jing Zhou and Jingren Zhou and Junyang Lin and Kai Dang and Keqin Bao and Kexin Yang and Le Yu and Lianghao Deng and Mei Li and Mingfeng Xue and Mingze Li and Pei Zhang and Peng Wang and Qin Zhu and Rui Men and Ruize Gao and Shixuan Liu and Shuang Luo and Tianhao Li and Tianyi Tang and Wenbiao Yin and Xingzhang Ren and Xinyu Wang and Xinyu Zhang and Xuancheng Ren and Yang Fan and Yang Su and Yichang Zhang and Yinger Zhang and Yu Wan and Yuqiong Liu and Zekun Wang and Zeyu Cui and Zhenru Zhang and Zhipeng Zhou and Zihan Qiu},
      year={2025},
      eprint={2505.09388},
      archivePrefix={arXiv},
      primaryClass={cs.CL},
      url={https://arxiv.org/abs/2505.09388}, 
}

@misc{bilal2024generatingzeroshotabstractiveexplanations,
      title={Generating Zero-shot Abstractive Explanations for Rumour Verification}, 
      author={Iman Munire Bilal and Preslav Nakov and Rob Procter and Maria Liakata},
      year={2024},
      eprint={2401.12713},
      archivePrefix={arXiv},
      primaryClass={cs.CL},
      url={https://arxiv.org/abs/2401.12713}, 
}

@Article{info13100500,
AUTHOR = {Jolly, Shailza and Atanasova, Pepa and Augenstein, Isabelle},
TITLE = {Generating Fluent Fact Checking Explanations with Unsupervised Post-Editing},
JOURNAL = {Information},
VOLUME = {13},
YEAR = {2022},
NUMBER = {10},
ARTICLE-NUMBER = {500},
URL = {https://www.mdpi.com/2078-2489/13/10/500},
ISSN = {2078-2489},
DOI = {10.3390/info13100500}
}

@inproceedings{atanasova-etal-2020-generating-fact,
    title = "Generating Fact Checking Explanations",
    author = "Atanasova, Pepa  and
      Simonsen, Jakob Grue  and
      Lioma, Christina  and
      Augenstein, Isabelle",
    editor = "Jurafsky, Dan  and
      Chai, Joyce  and
      Schluter, Natalie  and
      Tetreault, Joel",
    booktitle = "Proceedings of the 58th Annual Meeting of the Association for Computational Linguistics",
    month = jul,
    year = "2020",
    address = "Online",
    publisher = "Association for Computational Linguistics",
    url = "https://aclanthology.org/2020.acl-main.656/",
    doi = "10.18653/v1/2020.acl-main.656",
    pages = "7352--7364"
}

@INPROCEEDINGS{5597285,
  author={Brodersen, Kay Henning and Ong, Cheng Soon and Stephan, Klaas Enno and Buhmann, Joachim M.},
  booktitle={2010 20th International Conference on Pattern Recognition}, 
  title={The Balanced Accuracy and Its Posterior Distribution}, 
  year={2010},
  volume={},
  number={},
  pages={3121-3124},
  doi={10.1109/ICPR.2010.764}}

@article{zubiaga-etal-2016-pheme,
    doi = {10.1371/journal.pone.0150989},
    author = {Zubiaga, Arkaitz AND Liakata, Maria AND Procter, Rob AND Wong Sak Hoi, Geraldine AND Tolmie, Peter},
    journal = {PLOS ONE},
    publisher = {Public Library of Science},
    title = {Analysing How People Orient to and Spread Rumours in Social Media by Looking at Conversational Threads},
    year = {2016},
    month = {03},
    volume = {11},
    url = {https://doi.org/10.1371/journal.pone.0150989},
    pages = {1-29},
    number = {3},
}

\appendix

\section{Appendix}
\label{sec:appendix}

\section{Guidelines}
\label{app:guide}

Table \ref{tab:complex_guidelines} reports annotation guidelines for curating our dataset.

\begin{table*}[htb!]
\centering
\begin{tabular}{|p{2.0\columnwidth}|}
\hline 
Read the complex claim, its associated sub-claims, and the corresponding news articles. You will verify the veracity of each sub-claim based on the evidence found in the articles. \\ 

The complex claim will be broken down into **sub-claims**, each of which represents an atomic factual assertion. Your task is to highlight the portions of text within each news article that support or oppose the sub-claims. \\ 

\textbf{Highlighting Instructions:} \\ 
- Use a unique color for each sub-claim (e.g., blue for sub-claim 1, green for sub-claim 2, etc.). \\ 
- Highlight all relevant supporting or opposing text spans in the articles using the sub-claim’s color. \\
- If a text span is relevant to multiple sub-claims, split the highlight so each half uses the appropriate color. \\
- Irrelevant news articles or parts of articles should be deleted. \\ 
- If a single piece of evidence applies to multiple sub-claims, you may duplicate the evidence for each applicable sub-claim and highlight it accordingly. \\ 

\textbf{Veracity Annotation Rules:} \\ 
- \textbf{Unverified:} Choose this if (a) no evidence is found for the sub-claim or (b) the evidence found is contradictory (both support and oppose). All sub-claims start as unverified. \\ 
- \textbf{True:} Choose this if there is at least one highlight in at least one article that supports the sub-claim. \\
- \textbf{False:} Choose this if there is at least one highlight in at least one article that opposes the sub-claim. \\ 

\textbf{Example 1.} \\ 

\textbf{Complex Claim:} The Prime Minister announced a lockdown and curfew due to rising COVID-19 cases. \\ 

\textbf{sub-claim 1 (blue):} The Prime Minister announced a nationwide curfew. \\

\textbf{sub-claim 2 (green):} The curfew was implemented in response to rising COVID-19 cases. \\ 

\textbf{Evidence (Article Excerpt):} \\
1. "In a televised address, the Prime Minister declared a nationwide curfew starting at 8 PM." \\
2. "The government stated that the measures were necessary as daily COVID-19 cases reached record highs." \\ 

\textbf{Highlighting:} \\
- Highlight sentence 1 in blue (supports sub-claim 1). \\
- Highlight sentence 2 in green (supports sub-claim 2). \\ 

\textbf{Veracity:} \\
sub-claim 1: \textbf{True} \hspace{2em} sub-claim 2: \textbf{True} \\

\textbf{Example 2.} \\ 

\textbf{Complex Claim:} The recent protests were peaceful and did not involve any damage to public property. \\ \\

\textbf{sub-claim 1 (blue):} The protests were peaceful. \\
\textbf{sub-claim 2 (green):} No damage to public property occurred during the protests. \\ 

\textbf{Evidence (Article Excerpt):} \\
1. "Most protestors remained calm, though some clashes with police were reported late in the evening." \\
2. "Several reports emerged about broken windows and damaged street signs following the march." \\ 

\textbf{Highlighting:} \\
- Highlight sentence 1 in blue (partially opposes sub-claim 1). \\
- Highlight sentence 2 in green (opposes sub-claim 2). \\ 

\textbf{Veracity:} \\
sub-claim 1: \textbf{Unverified} \hspace{2em} sub-claim 2: \textbf{False} \\

\hline
\end{tabular}
\caption{Annotation guidelines for complex claim verification. Annotators identify and highlight evidence supporting or opposing each sub-claim, and label the sub-claim’s veracity accordingly.}
\label{tab:complex_guidelines}
\end{table*}

\section{Example}
\label{app:exp}
We provide an example in Table~\ref{tab:example_1} and Table~\ref{tab:example_2}.

\begin{table*}[htb!]
\centering
\begin{tabular}{|p{2.0\columnwidth}|}
\hline 
\textbf{Claim:} BREAKING: A \#Lufthansa \#Germanwings Airbus jet carrying 148 people has crashed in southern \#France, en route from \#Barcelona to \#Duesseldorf\\
\textbf{Veracity label:} T\\
\textbf{Full Evidence set:} France’s junior transport minister said there were “no survivors” from the crash of the Germanwings Airbus A320, a low-cost subsidiary of Lufthansa, in a remote part of the Alps that is extremely difficult to access. A Germanwings passenger jet carrying at least 150 people crashed in a snowy, remote section of the French Alps, sounding like an avalanche as it scattered pulverized debris across the mountain. Debris from crashed Germanwings Airbus A320 are seen in the mountains, near Seyne -les-Alpes, March 24, 2015.\\
Germanwings flight that crashed in a remote part of southern France today, according to a spokesperson for the city of Haltern, Germany The plane was en route from Barcelona, Spain, to Dusseldorf, Germany, when it crashed in the Alps, near the town of Digne , according to the airline. As emergency workers suspended recovery operations for the night, there was still little clue as to what caused the crash of a Germanwings Airbus A320 in a remote area of the French Alps on Tuesday, believed to have killed all 150 on board.\\
An Airbus operated by Lufthansa's Germanwings budget airline crashed into a mountainside in the French Alps on Tuesday, killing all 150 people on board including 16 schoolchildren. One of the plane's black box recorders has been found at the crash site, about 100 km north of the Riviera city of Nice, and will be examined immediately, France's interior minister said An Airbus A320 operated by the Germanwings flight 9525, a budget subsidiary of German carrier Lufthansa, crashed Tuesday in the French Alps, apparently killing all 150 people on board, including 16 high school students from Germany.\\
Airbus operated by Lufthansa's Germanwings budget airline crashed in a remote snowy area of the French Alps on Tuesday. An Airbus operated by Lufthansa's Germanwings budget airline crashed into a mountainside in the French Alps on Tuesday, killing all 150 people on board including 16 schoolchildren. Airbus A320 operated by the Germanwings airline crashed in mountains in southern France.\\
A Germanwings airliner carrying 144 passengers and six crew has crashed in the French Alps. A German jetliner en route from Barcelona, Spain, to Düsseldorf, Germany, plunged from the sky on Tuesday and slammed into the French Alps, killing all 150 people on board. Helicopters and rescue personnel swarmed into the remote, rugged area in southeastern France after the crash but found no signs of life. An Airbus A320 operated by the Germanwings flight 9525, a budget subsidiary of German carrier Lufthansa, crashed Tuesday in the French Alps, apparently killing all 150 people on board, including 16 high school students from Germany.\\
\hline
\end{tabular}
\caption{Example of claim resolution using sub-claims. Claim and evidence only}
\label{tab:example_1}
\end{table*}

\begin{table*}[htb!]
\centering
\begin{tabular}{|p{2.0\columnwidth}|}
\hline 
\textbf{Subclaims 1.} A Lufthansa Germanwings Airbus jet has crashed.\\
\textbf{Subclaim evidence:} "France’s junior transport minister said there were “no survivors” from the crash of the Germanwings Airbus A320, a low-cost subsidiary of Lufthansa, in a remote part of the Alps that is extremely difficult to access."\\
\textbf{Veracity label:} T\\
\textbf{Subclaims 2.} The jet was carrying 148 people. \\
\textbf{Subclaim evidence:} A Germanwings airliner carrying 144 passengers and six crew has crashed in the French Alps.\\ 
\textbf{Veracity label:} F\\
\textbf{Subclaims 3.} The crash occurred in southern France.\\
\textbf{Subclaim evidence:} "Germanwings flight that crashed in a remote part of southern France today, according to a spokesperson for the city of Haltern, Germany The plane was en route from Barcelona, Spain, to Dusseldorf, Germany, when it crashed in the Alps, near the town of Digne , according to the airline."\\
\textbf{Veracity label:} T
\textbf{Subclaims 4.} The jet was en route from Barcelona to Duesseldorf.\\
\textbf{Subclaim evidence:} "A German jetliner en route from Barcelona, Spain, to Düsseldorf, Germany, plunged from the sky on Tuesday and slammed into the French Alps, killing all 150 people on board."\\
\textbf{Veracity label: T}\\
Here we can see that the claim is mostly true except for casuaty number. This claim was likely made very early during the incident. Therefore, establishing the truth of most of information here is more important. Hence, the T veracity label at claim level. We can see that, breaking the claim down to subclaims and resolving them at subclaim level gives us better control over the resolution of the claim veracity.\\
\hline
\end{tabular}
\caption{Example of claim resolution using sub-claims.Sub-claim and resolution logic}
\label{tab:example_2}
\end{table*}

\section{Prompts}
\label{app:prompt}

Tables \ref{tab:prompts1}, \ref{tab:prompts2} and \ref{tab:prompts3} report prompt templates used in our experimental settings. In all prompts the evidence tags are the same. However, the sub-claim level evidence and document level evidence are separate in the main dataset which we just index in our prompt function. 

\begin{table*}[htb!]
\small
\centering
\begin{tabular}{|p{0.95\textwidth}|}
\hline
\textbf{Vanilla Baseline Prompt} \\ \\
{\ttfamily\raggedright
You are a journalist who specializes in fact-checking. You will receive a main claim and a set of supporting evidence. Your task is to determine the veracity of the claim based on the entire body of evidence provided, and then write a clear, concise explanation of your reasoning.

Guidelines::

1. Evaluate the relevance, credibility, and consistency of the evidence.

2. If the evidence directly supports the claim, the claim should be labeled True.

3. If the evidence clearly contradicts the claim, the claim should be labeled False.

4. If some evidence is missing or unverifiable, weigh the strength of the remaining evidence---the claim may still be True if key parts are well-supported.

5. Avoid requiring perfection. A claim can be True even if some evidence is incomplete, as long as the overall support is strong and consistent.

6. Use balanced, holistic reasoning. Consider the collective impact of all evidence when deciding veracity.

Formatting:

The claim will be inside <|Claim start|> and <|Claim end|> tags, and

the evidence will be inside <[Evidence start]> and <[Evidence end]> tags.

After evaluating the evidence, provide a short explanation followed by a final veracity label.

The label should be either T (True) or F (False).

The output should be in the following format:

Output format
<|journalist|> <explanation here>\\
Veracity: T/F.\par}
\\ \hline
\end{tabular}
\caption{Prompt templates used in our experiments. Vanilla operates on claim-level evidence only.}
\label{tab:prompts1}
\end{table*}

\begin{table*}[htb!]
\small
\centering
\begin{tabular}{|p{0.95\textwidth}|}
\hline
\textbf{Vanilla Baseline Prompt} \\ \\
{\ttfamily\raggedright
You are a journalist who specializes in fact-checking. You have been given a claim, its associated sub-claims, sub-claim veracity and supporting evidences. Your task is to determine the veracity of the overall claim based on the sub-claims and the provided evidence.\\

Guidelines:\\
1. Examine the evidence to confirm or challenge each sub-claim.\\
2. Treat sub-claims as hypotheses that must be evaluated using evidence.\\
3. Weigh sub-claims by their importance to the main claim.\\
4. Use holistic reasoning across all sub-claims and evidence.\\

Formatting:
Claim: \texttt{<|Claim start|>} ... \texttt{<|Claim end|>}\\
Subclaim: \texttt{<[Subclaim start]>} ... \texttt{<[Subclaim end]>}\\
Evidence: \texttt{<[Evidence start]>} ... \texttt{<[Evidence end]>}\\

Output format:\\
\texttt{<|journalist|> <brief explanation>}\\
\texttt{Veracity: T/F.}\\

The output should be in the following format:\\
<|journalist|> <explanation here>\\
Veracity: T/F.\par}
\\ \hline
\end{tabular}
\caption{Prompt templates used in our experiments. Oracle SRE uses claims with sub-claims, sub-claim veracity and claim level evidence.}
\label{tab:prompts2}
\end{table*}

\begin{table*}[htb!]
\small
\centering
\begin{tabular}{|p{0.95\textwidth}|}
\hline
\textbf{Vanilla Baseline Prompt} \\ \\
{\ttfamily\raggedright
You are a journalist who specializes in fact-checking. You have been given a claim, its associated sub-claims, and supporting evidences. Each sub-claim comes with a veracity label (T for True, F for False, U for Unverified). Your task is to determine the veracity of the overall claim based on the sub-claims and the provided evidence.\\

Guidelines:\\
1. Examine the evidence to confirm or challenge each sub-claim.\\
2. Do not blindly trust sub-claim veracity labels; treat them as hypotheses.\\
3. Weigh sub-claims by importance to the main claim.\\
4. Use holistic reasoning across sub-claims and evidence.\\

Formatting:\\
Claim: \texttt{<|Claim start|>} ... \texttt{<|Claim end|>}\\
Sub-claim: \texttt{<[Sub-claim start]>} ... \texttt{<[Sub-claim end]>}\\
Sub-claim veracity: \texttt{<[Sub-claim veracity start]>} ... \texttt{<[Sub-claim veracity end]>}\\
Evidence: \texttt{<[Evidence start]>} ... \texttt{<[Evidence end]>}\\

Output format:\\
<|journalist|> <brief explanation>\\
Veracity: T/F.\par}
\\ \hline
\end{tabular}
\caption{Prompt templates used in our experiments. Oracle SAE uses claims with sub-claims, sub-claim label and sub-claim aligned evidence.}
\label{tab:prompts3}
\end{table*}

\section{Models}
\label{app:llm_models}
For claim-level verification, we initially selected Llama 3.1 8B-instruct, due to its widespread use. However, LLama failed to perform the task. We found similar issues with Phi4-reasoning 14B. Our final selection was the Qwen3-14B reasoning model due to its successful handling of long context. We set top\_p (0.75) and top\_k (50) decoding with temperature of 0.3 and max new token of 8172. The input context was set at 16384 for SAE and 40960 for SRE. We also truncated anything that went above 40960 token length for SRE
\FloatBarrier
\clearpage
\section{Graph Neural Network}
\label{app:gnn}
The GNN uses a a bipartite graph to perform sub-claim veracity prediction The bipartite graph comprises of two distinct node sets: tweet nodes ($\mathbf{x}_t \in \mathbb{R}^{n \times 512}$) and evidence nodes ($\mathbf{x}_e \in \mathbb{R}^{m \times 128}$), interconnected via dedicated edges. The Twitter subgraph retains the original thread structure, while the evidence graph is built manually with the claim being the root node and the evidence articles spread out in a circular structure around the central node. Lastly, an additional bipartite connection explicitly links the tweet root node to the evidence root node. We created the node embeddings using a Roberta-base model. Finally, we process these node features through separate layers. Our GNN architecture processes these node features through separate layers: tweets go through mean-aggregation SAGEConv layers, followed by a graph attention layer (GATv2Conv). Evidence nodes similarly pass through distinct SAGEConv layers and a subsequent GATv2Conv attention layer. The embeddings are then aggregated using global max-pooling and combined via multi-head attention, culminating in a linear classification layer to predict claim veracity. The training was done with leave-one-event-out validation  and we used a learning rate of $5e^{-5}$, weight decay of $1e^{-4}$ and a batch size of $20$ .\label{app:Gnn}
\begin{figure*}[!t]
\centering
\includegraphics[width=\linewidth]{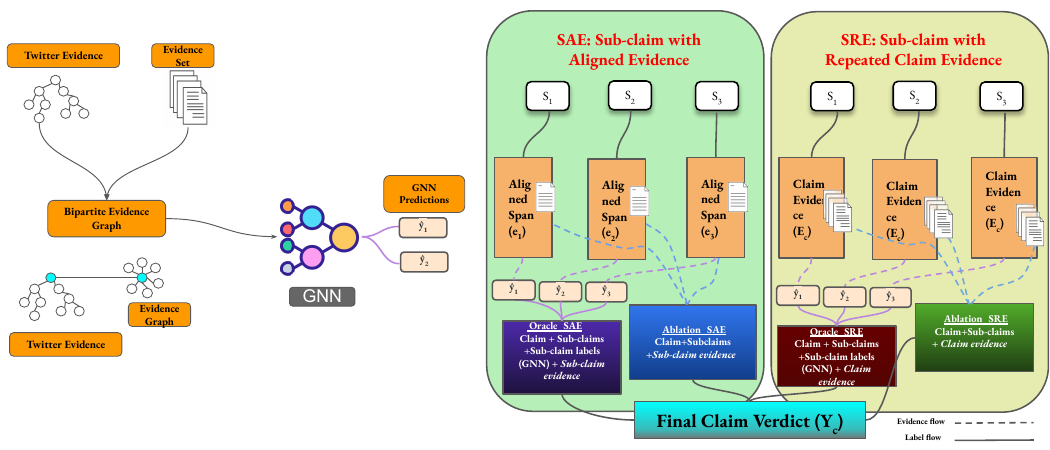}
\caption{\label{fig:gnn} Graph Neural network classification pipeline.}
\end{figure*}

\end{document}